\documentclass[11pt]{article}

\usepackage[preprint]{acl}

\usepackage{times}
\usepackage{latexsym}
\usepackage[section]{placeins} %
\usepackage{stfloats} %
\usepackage{capt-of} %

\usepackage[T1]{fontenc}

\usepackage[utf8]{inputenc}

\usepackage{microtype}

\usepackage{inconsolata}

\usepackage{graphicx}

\usepackage{url}            %
\usepackage{booktabs}       %
\usepackage{amsfonts}       %
\usepackage{nicefrac}       %
\usepackage{microtype}      %
\usepackage{xcolor}         %
\usepackage{amsmath}
\usepackage{amssymb}
\usepackage{pifont}
\usepackage{amssymb}
\usepackage{graphicx}
\usepackage{subcaption}
\usepackage{array}
\newcolumntype{C}[1]{>{\centering\arraybackslash}p{#1}}
\newcommand{\cmark}{\ding{51}} 
\newcommand{\xmark}{\ding{55}} 

\usepackage[english, status=draft, margin=false, inline=true]{fixme}
\fxusetheme{color}
\FXRegisterAuthor{jf}{ajf}{\color{red}JF}
\FXRegisterAuthor{mf}{mfa}{\color{blue}MFA}

\usepackage{enumitem}
\usepackage{multirow}

\newcommand{\githuburl}{\url{https://github.com/slinusc/bench360}}

\newcommand{\rmspace}{\vspace{-2ex}}
\newcommand{\added}[1]{{\color{black}#1}}

\title{Bench360: Benchmarking Local LLM Inference from 360 Degrees}

\author{Linus Stuhlmann \\
Zurich University of Applied Sciences\\
\texttt{linus.stuhlmann@zhaw.ch} \\
\And
Mauricio Fadel Argerich \\
Universidad Politécnica de Madrid \\
\texttt{mauricio.fadel@alumnos.upm.es} \\
\AND
Jonathan Fürst \\
Zurich University of Applied Sciences\\
\texttt{jonathan.fuerst@zhaw.ch}
}

\begin{document}

\maketitle

\begin{abstract}
Running LLMs locally has become increasingly common, but users face a complex design space across models, quantization levels, inference engines, and serving scenarios. Existing inference benchmarks are fragmented and focus on isolated goals, offering little guidance for practical deployments. We present Bench360, a framework for evaluating local LLM inference across tasks, usage patterns, and system metrics in one place. Bench360 supports custom tasks, integrates multiple inference engines and quantization formats, and reports both task quality and system behavior (latency, throughput, energy, startup time). We demonstrate it on four NLP tasks across three GPUs and four engines, showing how 
design choices shape efficiency and output quality. Results confirm that tradeoffs are substantial and configuration choices depend on specific workloads and constraints. There is no universal best option, underscoring the need for comprehensive, deployment-oriented benchmarks.

\end{abstract}

\section{Introduction}

Recent advances in quantization techniques (e.g., BitsAndBytes~\cite{dettmers2022llmint8, dettmers2023qlora} AutoAWQ~\cite{lin2023awq}, GPTQ~\cite{frantar2022gptq}), lightweight model architectures (e.g., Llama~\cite{touvron2023llama}, Gemma~\cite{team2024gemma}, Qwen~\cite{yang2024qwen2} and Mistral~\cite{mistral2023}), and inference frameworks (e.g., \added{\texttt{HuggingFace TGI}~\cite{huggingface_tgi_v3}, \texttt{vLLM}~\cite{kwon2023efficient}, \texttt{SGLang}~\cite{zheng2024sglang}, LMDeploy~\cite{2023lmdeploy, zhang2025efficient}}) have made local LLM deployments feasible for \textit{various tasks} across \textit{different deployment scenarios}. 
Besides task-specific functional quality metrics (e.g., accuracy, F1), local LLM deployments have varied non-functional requirements related to system performance (e.g., latency/throughput) and resource usage (e.g. energy).
All this constitutes a complex \textit{LLM inference design space} in which users deploying local models face a critical question: \textit{Which combination of model, quantization, and inference framework offers the best trade-off between computing cost, energy consumption, and output quality for my specific task and deployment scenario?}

While users can manually implement specific configurations and iteratively find a working solution through trial and error, this requires considerable time and effort and can still miss the best configuration. Users need to monitor relevant system performance and resource metrics alongside task-specific metrics like accuracy. What is missing is a benchmarking framework that allows them to: (1) select a set of potentially promising LLMs; (2) define the task with relevant task-specific metrics; (3) define their specific workload scenario; and (4) choose inference engines to evaluate. This framework should provide standardized evaluation across configurations and deliver comprehensive performance data to help users identify suitable configurations for their specific requirements. This is exactly what we propose with \textit{Bench360—Benchmarking Local LLM Inference from 360°}: \githuburl. Our framework systematically evaluates configurations across models, inference engines, quantization schemes, and deployment scenarios, providing users with actionable insights to make informed deployment decisions rather than relying on costly trial-and-error approaches.

While many benchmarks have been presented to evaluate large language models---such as MLPerf Inference~\cite{reddi2019mlperf}, Holistic Evaluation of Language Models (HELM)~\cite{liang2023holistic}, and the HuggingFace Open LLM Leaderboard~\cite{open-llm-leaderboard}---most of them focus on cloud-scale deployment, model accuracy, or leaderboard-style comparison. They often lack support for key deployment metrics such as energy efficiency, cold start and latency---all of which are critical for realistic on-premise deployment in constrained environments. Furthermore, many existing benchmarks are tightly coupled to specific APIs or evaluation scenarios, limiting their adaptability (see Section~\ref{sec:related} for a detailed discussion).

In contrast, %
Bench360 is tailored to the underexplored but increasingly important setting of local LLM deployment for specific tasks, emphasizing supporting practical deployment issues.
Overall, Bench360 provides the following key features:

\begin{itemize}

    \item \textbf{System and Task Metrics.} Bench360 combines system and task-specific metrics in one single framework. We integrate not only common efficiency metrics, including latency, throughput, memory usage, cold-start behavior, concurrency resilience and energy consumption, but also task-specific quality metrics (e.g., accuracy, F1, ROUGE) that users can define freely.
    \textit{This enables the selection of a configuration based on individual trade-offs.}
    
    \item \textbf{Multiple Inference Frameworks and Quantizations.} Bench360 supports multiple LLM model families, quantization schemes, and inference frameworks, enabling a direct comparison across popular frameworks such as \added{\texttt{HuggingFace TGI}, \texttt{vLLM}, \texttt{SGLang}, or \texttt{LMDeploy}}. \emph{This enables users to find the best combination of model architecture, quantization and inference framework.}
   
    \item \textbf{Custom Task and Inference Scenario Definition.} %
    Bench360 comes with plug-and-play support for new tasks and model configurations, designed to enable reproducible experiments and continuous benchmarking. \textit{This allows users to run Bench360 not only for pre-defined tasks, but lets them easily define and benchmark their own.}
    
\end{itemize}

\section{Related Work}
\label{sec:related}

\begin{table*}[hbt]
\centering
\scriptsize
\caption{Overview of core capabilities in LLM benchmarks compared to Bench360}
\label{tab:benchmarks-core}
\resizebox{1\hsize}{!}{
\begin{tabular} %
{C{5cm} C{0.7cm} C{2cm} C{2cm} C{2cm} C{1.2cm} C{1.2cm} C{1.2cm}}
\toprule
\textbf{Benchmark} &
\multirow{3}{*}{\textbf{Local}}
& \multicolumn{3}{c}{\textbf{Supported System Metrics}} & \multicolumn{3}{c}{\textbf{Custom Benchmark Definition}} \\
\cmidrule(r){3-5} \cmidrule(r){6-8}
 & & \textbf{Performance} & \textbf{Resources} & \textbf{Deployment} & \textbf{Tasks} & \textbf{Quantization} & \textbf{Scenarios} \\
\midrule
Open LLM Lead. \cite{open-llm-leaderboard} & \textcolor{red}{\xmark} & \textcolor{red}{\xmark} & CO2 emissions & \textcolor{red}{\xmark}  & \textcolor{red}{\xmark} & \textcolor{red}{\xmark} & \textcolor{red}{\xmark} \\
Edge LLM Lead. \cite{edge-llm-leaderboard} & \textcolor{teal}{\cmark} & Latency %
& Model size & \textcolor{red}{\xmark}  & \textcolor{red}{\xmark} & \textcolor{red}{\xmark} & \textcolor{red}{\xmark} \\
LLMPerf \cite{llmperf2024} & \textcolor{red}{\xmark} & Latency %
& Energy & \textcolor{red}{\xmark} & \textcolor{red}{\xmark} & \textcolor{red}{\xmark} & \textcolor{red}{\xmark} \\
Vidur-Bench \cite{agrawal2024vidur} & \textcolor{teal}{\cmark} & Latency, %
TPS, Memory usage & MFU, KV Cache Util. & \textcolor{red}{\xmark} & \textcolor{red}{\xmark} & \textcolor{teal}{\cmark} & \textcolor{red}{\xmark} \\
MLPerf Inference \cite{tschand2024mlperf} & \textcolor{teal}{\cmark} & Latency, TPS %
& Energy & \textcolor{red}{\xmark} & \textcolor{red}{\xmark} & \textcolor{teal}{\cmark} & \textcolor{teal}{\cmark} \\
LLM-Inference-Bench~\cite{chitty2024llm} & \textcolor{teal}{\cmark} & Latency %
& Energy & \textcolor{red}{\xmark}  & \textcolor{teal}{\cmark} & \textcolor{teal}{\cmark} & \textcolor{red}{\xmark} \\ 
ML.ENERGY \cite{chung2025ml} & \textcolor{teal}{\cmark} & Latency 
& Energy  & \textcolor{red}{\xmark}  & \textcolor{red}{\xmark} & \textcolor{red}{\xmark} & \textcolor{red}{\xmark} \\ \midrule
\textbf{Bench360 (ours)} & \textcolor{teal}{\cmark} & Latency, %
Peak Memory usage, GPU util. & Energy, Cost per query, Model size, Overhead & Cold start time, OOM resil., Multi-user-concurrency & \textcolor{teal}{\cmark} & \textcolor{teal}{\cmark} & \textcolor{teal}{\cmark} \\
\bottomrule
\end{tabular}
}
\rmspace
\end{table*}

The widespread adoption of LLMs has led to a growing interest in benchmarking their inference performance. LLMeBench~\cite{dalvi2023llmebench} enables the evaluation of the accuracy of LLMs on different NLP tasks, but does not provide any efficiency metrics.
The Open LLM Leaderboard~\cite{open-llm-leaderboard} by Hugging Face focuses on evaluating the task-specific accuracy of models, but only reports the estimated carbon emissions during their evaluation. The Edge LLM Leaderboard~\cite{edge-llm-leaderboard} further includes system metrics such as time-to-first-token (TTFT) and inter-token-latency. LLMPerf~\cite{llmperf2024} evaluates the output quality via correctness and system metrics via a load test, but does not evaluate local deployments, only LLM APIs. 
Vidur-Bench~\cite{agrawal2024vidur} is an LLM benchmark for evaluating the performance of LLMs through workload patterns (tasks), scheduling, batching, and routing policies, and serving frameworks. Vidur-Bench is part of an LLM simulation framework, and new models require several steps that make it not easy to evaluate and compare multiple LLMs. \textit{Notably, all above benchmarks do not measure the energy consumption of LLMs.}
MLPerf~\cite{mattson2020mlperf}, a joint effort between academia and industry, develops a benchmark suite to evaluate the training and inference efficiency, as well as the quality, safety and energy of ML systems. MLPerf includes various benchmarks for training and inference in different domains (e.g, language, computer vision, and recommender systems).
Reference implementations are provided for multiple models in four different scenarios (single stream, multiple stream, server, and offline).
Unlike the previous benchmarks, MLPerf considers energy consumption through the MLPerf Power workgroup that measures energy consumption while running the MLPerf benchmarks through the use of a physical power meter---currently, software measurements are not accepted for submission~\cite{tschand2024mlperf}. 
LLM-Inference-Bench~\cite{chitty2024llm} is designed to evaluate the inference performance of LLMs in the Llama family and derivatives, such as Gemma, and Qwen, across multiple frameworks and hardware. However, LLM-Inference-Bench is task-agnostic, using only perplexity to evaluate the quality of LLMs, which is not a sufficient metric to approximate task-specific metrics~\cite{nimah2023nlg}.
\added{Last, ML.ENERGY~\cite{chung2025ml} addresses the automated energy optimization for large-scale LLM inference, focusing on Pareto-frontier-based configuration tuning for high-end data center GPUs.}
In contrast to these system-oriented benchmarks, Bench360 streamlines the evaluation of LLMs across tasks and scenarios on a given deployment environment.
Table~\ref{tab:benchmarks-core} summarizes the differences of Bench360 compared to existing works. \textit{Compared to all existing works we cover the entire local deployment stack including a wide range of system metrics and customizability of task, quantization and scenario.}

\section{Bench360---Benchmarking Local LLMs from 360\textdegree}
\label{sec:design}

Bench360 comprises four components: (1) a \textbf{task engine} that standardizes input/output for tasks such as summarization, QA, and Text-to-SQL; (2) a \textbf{workload controller} that simulates usage scenarios (single-stream, batch, server) with configurable query patterns; (3) a \textbf{backend abstraction layer} for executing models via different inference engines \added{(currently, HuggingFace TGI, vLLM, SGLang and LMDeploy)}; and (4) a \textbf{metrics collector} that captures system metrics such as latency, throughput, memory usage, energy consumption, and task-specific metrics.

\begin{figure*}[ht]
\centering
\includegraphics[width=0.8\linewidth]{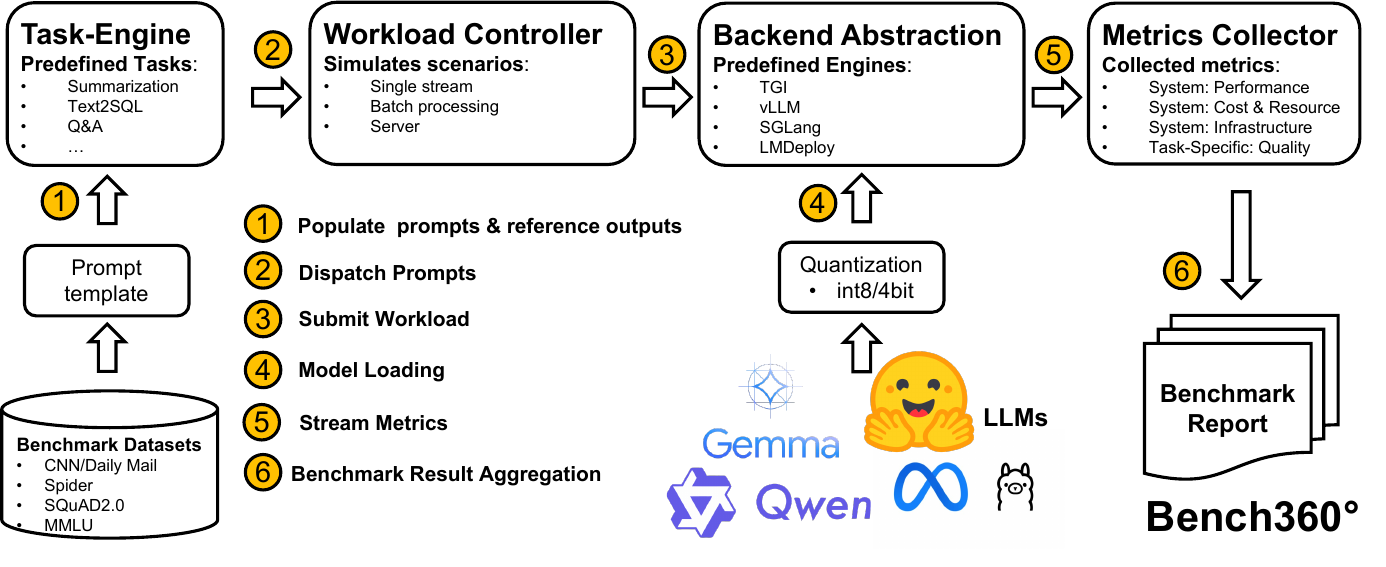}
\caption{\textbf{Bench360.}  
Datasets plus a prompt template yield task-specific prompts that enter the \emph{Task-Engine} (1) and pass to the \emph{Workload Controller} (2), which simulates single, batch, or server traffic before handing the workload to the \emph{Backend Abstraction} (3).  
The backend loads a quantized LLM and executes inference on the selected engines (4); generated text and runtime traces stream into the \emph{Metrics Collector} (5), which aggregates system and task metrics into a final benchmark report (6).  
}
\label{fig:bench360-overview}
\rmspace
\end{figure*}

Bench360 provides a command-line interface and configuration-based workflow. Here, users configure the benchmark via a YAML configuration file (Table~\ref{tab:config} in Appendix) without modifying code. This file specifies the task type, dataset, evaluation metrics, model configuration, backend engine, and quantization format.
For advanced use, developers can extend the framework by implementing new tasks, inference engines, or workload generators using the modular API defined in the benchmark core. \added{Models can be retrieved directly from the Hugging Face Hub in the required format, including quantized variants, provided that the chosen inference engine supports them.}

\subsection{Task Engine}
The task engine operates through a standardized plugin architecture where each task is linked to a dataset via a path or identifier (e.g., \texttt{mmlu}, \texttt{squad\_v2}, \texttt{cnn\_dailymail}, or custom local files). Each task implements the \texttt{BaseTask} interface with two core methods: \texttt{generate\_prompts()} which creates test prompts and extracts reference answers from the dataset, and \texttt{ quality\_metrics()} which computes task-specific evaluation scores by comparing generated outputs with reference answers. The benchmark automatically handles input formatting and output parsing for each task type. Tasks are dynamically loaded and instantiated based on configuration parameters, allowing the system to seamlessly switch between different evaluation scenarios. The task engine abstracts away dataset-specific preprocessing and metric computation, providing a unified interface that integrates with the broader benchmarking framework's inference engines and deployment scenarios.

\subsection{Workload Controller}%
Inspired by MLPerf~\cite{mattson2020mlperf}, the controller emulates three serving scenarios—\emph{single-stream}, \emph{batch}, and \emph{server}—each implemented by a unified event loop that advances in fixed time steps.
\textbf{Single-stream.} Exactly one prompt is kept ready at all times. Whenever the backend becomes idle, that prompt is sent, the loop blocks until the answer is returned, and the next prompt is loaded. No queue is formed.
\textbf{Batch.} All prompts are available before the timer starts. The controller accumulates a batch of size $b$ and submits it in one call. This repeats until all samples are processed.
\textbf{Server.} The server scenario simulates concurrent users making requests to the inference backend. Each user generates requests independently following a Poisson process with a predefined rate of queries per minute. The benchmark uses a thread pool to dispatch requests immediately to the backend as they arrive, there is no artificial queuing or request batching imposed by the benchmark itself. Any request queuing, batching, or scheduling behavior observed during execution is entirely determined by the backend's internal implementation and capacity constraints.

\subsection{Backend Abstraction}\label{sec:backend}
The \textbf{Backend Abstraction} layer provides a unified interface for multiple inference engines including Hugging Face TGI, vLLM, SGLang and LMDeploy. Each backend is containerized using Docker to ensure consistent deployment and isolation. At startup, the system launches the appropriate container with the requested model, loading from local storage or automatically downloading from Hugging Face Hub. The abstraction layer translates unified configuration parameters into each engine's native API format and dispatches requests through standardized OpenAI-compatible endpoints. This design allows seamless comparison between %
engines and quantization schemes without configuration changes, while new runtimes can be integrated by implementing the common backend interface.

\subsection{System Metrics Collector}

We categorize system metrics into \textit{Performance}, \textit{Resource}, and \textit{Deployment}
(Table~\ref{tab:system_evaluation_metrics}). \textbf{Performance.} \added{We adopt the full NVIDIA NIM metrics suite~\cite{nvidia_nim_metrics}:} We measure latency at different granularities including time-to-first-token (TTFT), time per output token (TPOT), and generation latency (GL). Throughput is tracked in tokens per second (TPS) and sentences per second (SPS). These metrics are recorded during generation using consistent timing hooks across all backends.
\textbf{Resources.} We track GPU utilization, memory footprint, CPU utilization, and RAM usage of the inference process using NVML and psutil telemetry. We focus on the GPU's energy consumption since it dominates LLM inference~\cite{argerich2024measuring} and measure it via NVML, shown to have errors below $5\%$~\cite{yang2024accurate} and being well-suited for comparisons across deployments. Model size and memory overhead are computed from disk footprint and runtime memory deltas. Energy efficiency is measured as energy per token and energy per sentence.

\textbf{Deployment.} We evaluate comprehensive deployment metrics including container startup time, model loading time, and cold-start latency. For server scenarios, we additionally measure queue time (request submission to processing delay), wait time (scheduling overhead), and end-to-end latency to capture real-world deployment behavior under concurrent load.

 \begin{table}[hb]
  \centering
  \caption{System Evaluation Metrics Overview}
  \resizebox{1.0\linewidth}{!}{%
  \begin{tabular}{llll}
  \toprule
  \textbf{Category} & \textbf{Aspect} & \textbf{Metric} & \textbf{Measurement Details} \\
  \midrule
  \textbf{Performance} & Latency & TTFT, TPOT, GL & Timing hooks \\
  & Throughput & TPS, SPS& Tokens/sentences/second \\
  & Server Latency & Queue time, wait time, E2E latency & Thread pool \\
  \midrule
  \textbf{Resources} & GPU Memory & Peak/average GPU memory (MB) & NVML memory monitoring \\
  & GPU Utilization & Peak/average GPU utilization (\%) & NVML utilization rates \\
  & CPU/RAM Usage & Peak/average CPU utilization, RAM (MB) & psutil process monitoring \\
  & Power & Peak/average power (W), total energy (Wh) & NVML power telemetry \\
  & Energy Efficiency & Energy per token/sentence (J) & Power × generation time \\
  & Memory Overhead & Runtime memory beyond model size & GPU memory delta \\
  \midrule
  \textbf{Deployment} & Cold Start & Container startup, TTFT, total coldstart & Launch to first token \\
  & Model Size & Disk footprint (MB) & Checkpoint file size \\
  \bottomrule\end{tabular}
  }
  \label{tab:system_evaluation_metrics}
  \rmspace
  \end{table}

\section{Experimental Evaluation}
\label{sec:evaluation}

While Bench360 is a general-purpose LLM inference benchmark, we rely on a set of research questions to guide us in our investigation and to show the unique benefits of our framework.
Specifically, for different LLM model sizes/families, for different hardware platforms and four tasks, we investigate the following questions.

\paragraph{RQ1: Memory-Constrained Deployment:} Given a strict 24GB VRAM budget (typical of mid-tier GPUs), how should memory be allocated most effectively? Is it better to run a smaller full-precision model (e.g., 7B FP16) or a larger quantized counterpart (e.g., 14B INT8, 24–32B INT4) in terms of overall effectiveness (quality) and efficiency (latency, energy per token)?
    
\paragraph{RQ2: Scenario-Focused Serving:} Across the three serving scenarios (single-stream responsiveness, offline batch throughput, and multi-user concurrency) which inference engine (vLLM, SGLang, LMDeploy, TGI) offers the best performance on a given GPU platform?

\paragraph{RQ3: System Efficiency:} Do inference engines differ in their energy consumption, and which engine–GPU combinations achieve the best efficiency relative to GPU utilization?

\subsection{Experimental Setup}
\label{subsec:experimental-setup}

\textbf{Tasks, datasets, and metrics.}  
We evaluate four tasks that reflect common LLM applications.  
First, Multitask Language Understanding covers reasoning, knowledge retrieval, and comprehension using the MMLU benchmark dataset~\cite{hendrycks2020measuring}, which is evaluated using accuracy.  
Second, Extractive Question Answering through SQuAD2.0~\cite{rajpurkar2018know}, where we measure F1 score. Third, we evaluate Summarization capabilities on the CNN/DailyMail dataset~\cite{see-etal-2017-get}, with ROUGE-L as the main metric. Fourth, we evaluate Text-to-SQL with the Spider dataset~\cite{yu2018spider}, using Execution Accuracy (EA) as the primary metric~\cite{furst2024evaluating, sivasubramaniam2024sm3, yu2018spider}.  

\textbf{LLMs, inference engines, and hardware.} We evaluate four open-weight LLM families (LLaMA-3, Gemma-2, Mistral, Qwen2.5) all within a 24GB VRAM constraint to reflect realistic local deployments.
For \emph{memory-constrained deployment}, we compare smaller FP16 models against larger GPTQ-quantized~\cite{gptq} variants from the same families: Gemma-2 (9B FP16 vs.\ 27B 
  INT4), Mistral (7B FP16 vs.\ 24B INT4), and Qwen2.5 (7B FP16, 14B INT8, 32B INT4) using vLLM across L4, A10, and A30 GPUs.
For \emph{scenario-focused serving}, we benchmark three 
  representative mid-size models (Mistral-7B-Instruct-v0.3, Qwen2.5-7B-Instruct, LLaMA-3.1-8B-Instruct) across four inference engines (vLLM, SGLang, TGI, LMDeploy) on all GPU tiers.
For 
  \emph{system efficiency}, we use the same three models and four engines across all three GPUs to isolate performance differences (Tables~\ref{tab:engine_comparison} and \ref{tab:gpu_comparison}).

\subsection{Memory-Constrained Deployment}
\label{subsec:vram-frontier}

A core deployment question is how to allocate a fixed GPU memory budget. Quantization provides a solution, as it enables larger models to fit into limited memory, but it typically comes at the cost of quality degradation~\cite{li2024evaluatingquantizedlargelanguage}. An important, under-investigated aspect for practical deployments is whether a larger quantized model can outperform a smaller FP16 counterpart under the same memory constraints. To answer this, we compare, for each model family, the largest FP16 model that fits within 24 GB against the largest quantized (INT8/INT4) variant.
Our evaluation spans tasks in general knowledge and reasoning (MMLU), extractive QA (SQuAD v2), summarization (CNN/DailyMail), and Text-to-SQL (Spider).

\begin{figure*}[htbp]
  \centering
  \includegraphics[width=0.80\linewidth]{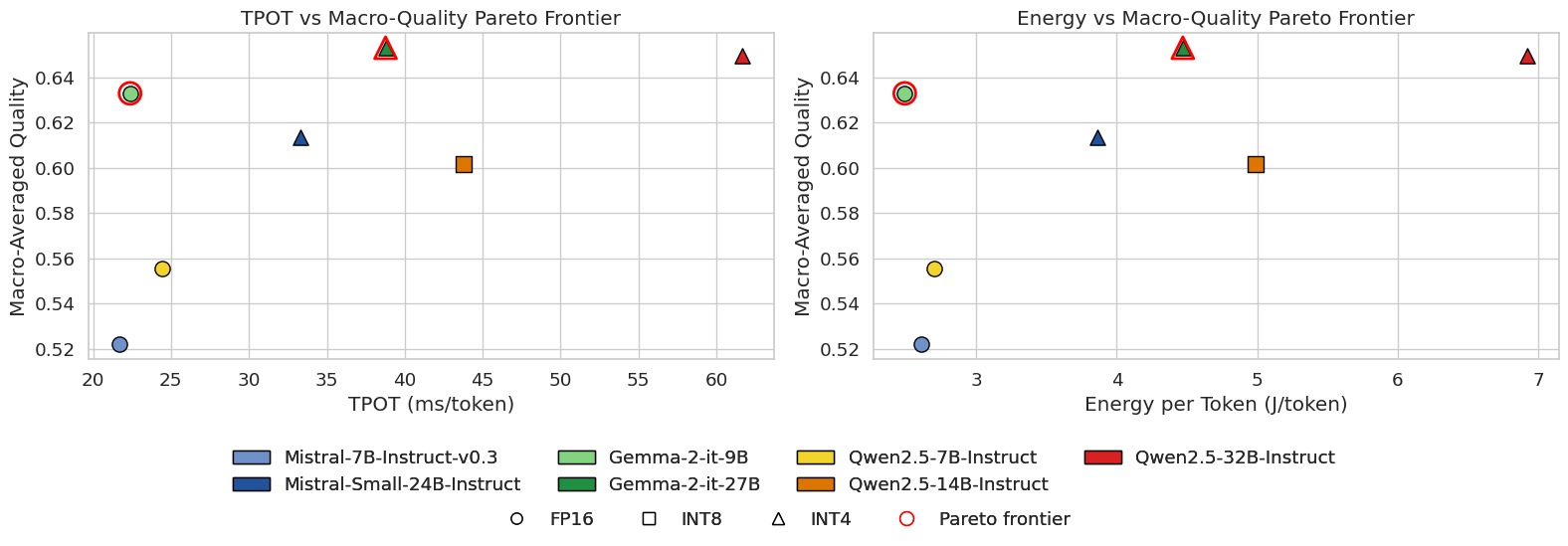}
  \caption{Latency–quality and energy–quality Pareto frontiers under a 24\,GB VRAM budget. Gemma-2 variants dominate both frontiers, while Qwen2.5 achieves the highest task quality at disproportionate system cost.}
  \label{fig:model_pareto}
    \rmspace
\end{figure*}

\begin{table}[htbp]
\centering
\footnotesize 
\setlength{\tabcolsep}{3.5pt} %

\begin{tabular}{@{}ll r r r@{}}
 & & \textbf{Performance} & \textbf{Latency} & \textbf{Efficiency} \\
\cmidrule(lr){3-3} \cmidrule(lr){4-4} \cmidrule(lr){5-5}
\textbf{Model} & \textbf{Q} & \textbf{Score} & \textbf{ms} & \textbf{J/Tok} \\

\midrule
\multicolumn{5}{l}{\textit{MMLU (Accuracy)}} \\
\midrule
Gemma-2-9B & FP16 & 0.714 & \textbf{47.6} & \textbf{4.65} \\
Gemma-2-27B & INT4 & 0.742 & 125.5 & 12.71 \\
\addlinespace[0.3em]
Mistral-7B & FP16 & 0.574 & 89.6 & 9.03 \\
Mistral-24B & INT4 & 0.574 & 110.5 & 10.44 \\
\addlinespace[0.3em]
Qwen2.5-7B & FP16 & 0.688 & 189.7 & 17.86 \\
Qwen2.5-14B & INT8 & 0.769 & 266.2 & 26.14 \\
Qwen2.5-32B & INT4 & \textbf{0.794} & 336.9 & 32.93 \\

\midrule
\multicolumn{5}{l}{\textit{Squad v2 (F1 Score)}} \\
\midrule
Gemma-2-9B & FP16 & 0.903 & 36.5 & \textbf{4.12} \\
Gemma-2-27B & INT4 & 0.907 & 78.1 & 9.03 \\
\addlinespace[0.3em]
Mistral-7B & FP16 & 0.847 & \textbf{35.4} & 4.23 \\
Mistral-24B & INT4 & \textbf{0.916} & 58.8 & 6.82 \\
\addlinespace[0.3em]
Qwen2.5-7B & FP16 & 0.871 & 57.4 & 6.32 \\
Qwen2.5-14B & INT8 & 0.885 & 98.5 & 11.24 \\
Qwen2.5-32B & INT4 & 0.889 & 128.2 & 14.34 \\

\midrule
\multicolumn{5}{l}{\textit{CNN/DM (ROUGE-L)}} \\
\midrule
Gemma-2-9B & FP16 & 0.380 & 16.3 & 1.75 \\
Gemma-2-27B & INT4 & 0.395 & 18.2 & 2.09 \\
\addlinespace[0.3em]
Mistral-7B & FP16 & 0.384 & \textbf{7.0} & \textbf{0.81} \\
Mistral-24B & INT4 & \textbf{0.396} & 14.9 & 1.75 \\
\addlinespace[0.3em]
Qwen2.5-7B & FP16 & 0.231 & 7.2 & 0.82 \\
Qwen2.5-14B & INT8 & 0.370 & 11.7 & 1.35 \\
Qwen2.5-32B & INT4 & 0.378 & 32.5 & 3.68 \\

\midrule
\multicolumn{5}{l}{\textit{Spider (Exec. Acc.)}} \\
\midrule
Gemma-2-9B & FP16 & 0.616 & 14.0 & 1.61 \\
Gemma-2-27B & INT4 & 0.657 & 20.0 & 2.30 \\
\addlinespace[0.3em]
Mistral-7B & FP16 & 0.335 & 22.8 & 2.78 \\
Mistral-24B & INT4 & 0.528 & 26.2 & 3.02 \\
\addlinespace[0.3em]
Qwen2.5-7B & FP16 & 0.565 & \textbf{8.7} & \textbf{0.96} \\
Qwen2.5-14B & INT8 & 0.550 & 21.1 & 2.38 \\
Qwen2.5-32B & INT4 & \textbf{0.681} & 24.4 & 2.75 \\

\bottomrule
\end{tabular}
\caption{Performance, latency, and energy efficiency across tasks. \textbf{Performance} refers to the specific metric listed in each section header. \textbf{Latency} = TPOT (ms). \textbf{Efficiency} = Joules/Token.}
\label{tab:vram_ladder_minimal}
\rmspace
\end{table}

Table~\ref{tab:vram_ladder_minimal} and Fig.~\ref{fig:model_pareto} reveal three consistent patterns across tasks.  
(i) Quantization enables the use of large models but the higher number of operations per token still requires more computing. INT8/INT4 allows 2--4$\times$ larger models within a 24GB budget and often yields higher scores, yet TPOT and energy can rise by more than $+350\%$, especially for Qwen2.5-32B INT4, impacted by the higher number of operations needed. The roofline model shows that memory and computation are the two main bottlenecks in LLM inference~\cite{yuan2024llm}; while quantization has been shown to help in reducing memory and bandwidth bottlenecks in LLM inference~\cite{argerich2024measuring}, computation is not affected by quantization, thus its benefit is limited. Other techniques such as pruning could help to reduce the computation bottleneck~\cite{chitty2023survey, park2024inference}.
(ii) Efficiency is family-dependent. Gemma-2 achieves the most balanced trade-off: its 9B FP16 baseline already sits near the Pareto frontiers, and the 27B INT4 variant improves accuracy with moderate overheads. In contrast, Mistral-24B INT4 achieves large quality gains (up to $+57.8\%$ in SQL) but with notable latency and energy penalties, while Qwen2.5 maximizes accuracy at the steepest cost in both TPOT and J/Token.  
(iii) System cost depends more on architecture than on size alone. Smaller FP16 baselines like Gemma-9B and Mistral-7B remain more energy-efficient than their quantized large-family counterparts, indicating that activation compute and memory bottlenecks dominate runtime, even with compressed weights and activations. Only the Gemma-2 series consistently appears on both the latency--quality and energy--quality Pareto frontiers, showing that architectural design is as critical as model scale.

\paragraph{Answering RQ1}
Quantized large models can outperform smaller FP16 baselines within a fixed VRAM budget, but the trade-offs vary by family.  
Gemma-9B FP16 is best for latency-sensitive deployments, combining strong accuracy with the lowest TPOT and J/Token in its block.  
Gemma-27B INT4 provides the most balanced upgrade path, lifting accuracy without catastrophic overheads.  
Mistral-24B INT4 and Qwen2.5-32B INT4 achieve the highest task quality improvements (Table~\ref{tab:vram_ladder_minimal}) but at the steepest system cost, making them suitable for batch-serving or offline workloads where throughput matters more than latency.  
Overall, optimal deployment depends on jointly considering quantization level, architecture, and task requirements; 
\emph{choosing which family to scale often matters more than simply scaling further}.

\subsection{Scenario–Focused Serving}
\label{subsec:usecase-serving}

A core serving question is how to provision an inference stack for a fixed GPU type and inference scenario. We study whether inference frameworks deliver the responsiveness, throughput, and energy efficiency required by the three dominant deployment regimes—\emph{single-stream}, \emph{offline batch}, and \emph{multi-user concurrency}—and how these trade-offs vary across hardware. %

\subsubsection{Single-Stream Latency}
\label{subsec:discussion-single-stream}

\begin{figure*}[htbp]
  \centering
  \includegraphics[width=0.9\textwidth]{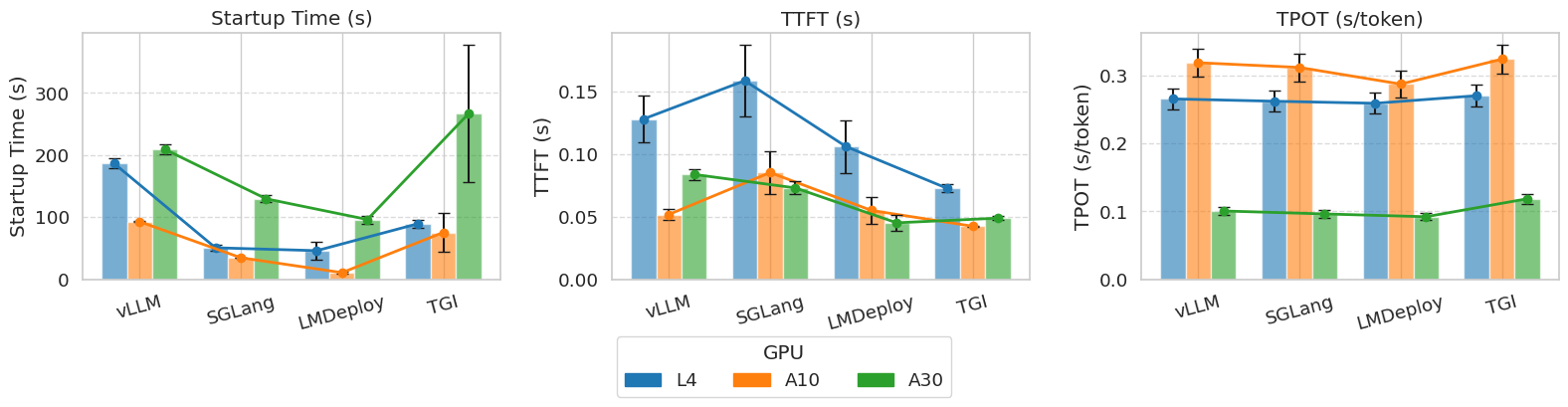}
  \caption{\textbf{Single-stream responsiveness.} Startup time, TTFT, and TPOT averaged over all tasks/models on L4, A10, and A30. Error bars indicate 95\% confidence intervals.}
  \label{fig:single}
    \rmspace
\end{figure*}

Figure~\ref{fig:single} summarises startup time, TTFT, and TPOT under single-stream conditions. LMDeploy exhibits the strongest end-to-end responsiveness overall: it starts fastest, produces the first token earliest, and sustains the lowest TPOT across devices. On A10 and L4, startup completes in roughly 15\,s and 40\,s, respectively, while A30 also benefits from markedly shorter cold-start delays than competing engines. By contrast, TGI shows protracted and highly variable startup, exceeding 260\,s on A30 with wide confidence intervals, which undermines interactive use. vLLM and SGLang start more quickly than TGI but do not match LMDeploy’s cold-start profile.

First-token experience follows a similar pattern. LMDeploy and TGI obtain the lowest TTFT on all three GPUs, with LMDeploy reaching about 45ms on A30 and maintaining tight run-to-run variability. vLLM and SGLang produce competitive but consistently higher TTFTs, especially on L4 where SGLang peaks near 160ms. Once generation is underway, TPOT is lowest on A30 for all engines due to higher memory bandwidth. LMDeploy again leads, recording $\sim$0.09\,s/token on L4 and $\sim$0.095\,s/token on A30, with vLLM and SGLang following closely and TGI lagging at both higher TPOT and higher variance.

\subsubsection{Batch Throughput}
\label{subsec:discussion-batching}

We scale the throughput evaluation under increasing batch sizes $B \in \{16, 32, 64, 128\}$. Table~\ref{tab:tps_vertical_gpu} reports offline throughput scaling as batch size increases. All engines benefit from larger batches, but the winner depends on the GPU tier. On mid-range hardware (L4, A10), \textbf{SGLang} delivers the highest TPS across all batch sizes, reaching 648.5\,TPS on A10 at $B{=}128$ and consistently outpacing vLLM, LMDeploy, and TGI. This suggests that SGLang’s batching and kernel execution are particularly effective under bandwidth and compute constraints. On the high-end A30, \textbf{vLLM} takes the lead, peaking at 964.4\,TPS for $B{=}128$, which points to runtime optimisations that better exploit tensor cores and high-bandwidth memory. LMDeploy and TGI scale less efficiently, especially on L4 and A10 where their TPS often remains below 200 at the largest batch sizes. Since all experiments use FP16 variants of Qwen2.5-7B, LLaMA-3.1-8B, and Mistral-7B, the observed differences chiefly reflect engine-level scheduling and execution efficiency rather than numerical compression.

\begin{table}[ht]
\centering
\resizebox{1.0\linewidth}{!}{%
\begin{tabular}{llcccc}
\toprule
\textbf{GPU} & \textbf{Batch Size} & \textbf{vLLM} & \textbf{SGLang} & \textbf{LMDeploy} & \textbf{TGI} \\
\midrule
\multirow{4}{*}{\textbf{L4}}
& 16  & 133.0 ± 5.6 & \textbf{140.7 ± 6.1} & 100.4 ± 4.1 & 92.1 ± 3.5 \\
& 32  & 190.6 ± 8.8 & \textbf{228.4 ± 11.1} & 137.4 ± 6.1 & 116.5 ± 4.8 \\
& 64  & 304.8 ± 15.2 & \textbf{364.3 ± 19.8} & 162.5 ± 8.0 & 135.3 ± 5.9 \\
& 128 & 401.9 ± 21.0 & \textbf{513.2 ± 29.4} & 198.6 ± 11.5 & 166.5 ± 8.8 \\
\midrule
\multirow{4}{*}{\textbf{A10}}
& 16  & 155.3 ± 7.2 & \textbf{173.9 ± 8.6} & 112.3 ± 4.7 & 96.2 ± 3.7 \\
& 32  & 253.1 ± 12.9 & \textbf{296.2 ± 16.3} & 142.8 ± 6.2 & 120.5 ± 4.7 \\
& 64  & 377.8 ± 21.4 & \textbf{463.0 ± 28.2} & 162.3 ± 7.6 & 139.6 ± 5.6 \\
& 128 & 497.8 ± 29.6 & \textbf{648.5 ± 41.4} & 176.5 ± 8.6 & 149.0 ± 6.1 \\
\midrule
\multirow{4}{*}{\textbf{A30}}
& 16  & \textbf{318.6 ± 12.7} & 218.5 ± 6.7 & 238.6 ± 9.5 & 173.2 ± 6.0 \\
& 32  & \textbf{442.6 ± 19.5} & 362.0 ± 12.2 & 312.2 ± 13.6 & 228.2 ± 8.4 \\
& 64  & \textbf{733.3 ± 34.8} & 574.4 ± 23.4 & 353.0 ± 16.6 & 246.1 ± 9.0 \\
& 128 & \textbf{964.4 ± 47.9} & 917.6 ± 43.6 & 378.2 ± 18.6 & 263.0 ± 9.8 \\
\bottomrule
\end{tabular}
}
\caption{\textbf{Offline batch throughput} (tokens/s ± 95\% CI) across batch sizes and GPUs.}
\label{tab:tps_vertical_gpu}
\rmspace
\end{table}

\subsubsection{Server Concurrency}
\label{subsec:discussion-concurrency}

We now turn to multi-user serving with short-to-medium prompts and outputs (320
tokens prompts and outputs 32 tokens output in average). For concurrency we use $C\!\in\!\{8,16,32,64\}$ simulated threads at 12\,rpm with Poisson arrivals and QoS (Quality of Service) thresholds of 2\,s (TTFT) and 6\,s (E2E). Throughput increases with concurrency until each engine-GPU pair reaches its saturation point, at which latency grows rapidly. Figure \ref{fig:server_concurrency_all} shows vLLM continues to scale up to $C{=}64$ on A30, while keeping TTFT and E2E close to their targets, whereas on A10 and L4, saturation typically occurs between $C{=}16$ and $C{=}32$ due to tighter %
budgets. Across devices, vLLM exhibits the most robust queueing and scheduling behaviour: it consistently satisfies QoS up to $C{=}32$ and remains usable at $C{=}64$ without severe tail latencies. SGLang performs well at light to moderate concurrency but breaches the TTFT threshold earlier on L4 as load increases. LMDeploy and TGI saturate sooner, with rising TTFT and E2E beyond $C{=}16$ in many settings, indicating less effective request interleaving and cache management. For chat backends and shared APIs where predictable latency under load is essential, our results favour vLLM.

\paragraph{Answering RQ2}
These results underline that \emph{the optimal inference engine strongly depends on the serving scenario}. LMDeploy’s short startup and low TTFT make it the most suitable choice when single-request responsiveness and frequent cold starts dominate the user experience, such as in CLI tools, on-demand inference, debugging loops, and serverless or edge deployments. SGLang shows clear advantages in offline batch settings on mid-tier GPUs (L4, A10), whereas vLLM dominates on upper-mid hardware (A30), where its runtime optimizations fully exploit memory bandwidth and tensor-core throughput. In multi-user concurrency, vLLM emerges as the most robust engine, consistently maintaining QoS under high load, while SGLang remains a solid second choice. By contrast, LMDeploy and TGI saturate early, making them less suitable for shared backends. Taken together, these findings imply that engine choice should be aligned with the intended deployment pattern: LMDeploy for cold-start responsiveness, SGLang for batch workloads on constrained devices, and vLLM for scalable multi-user serving, especially on high-bandwidth GPUs.

\subsection{System Efficiency}
\label{subsec:discussion-efficiency}

\begin{figure}[htbp]
  \centering
  \includegraphics[width=1.05\linewidth]{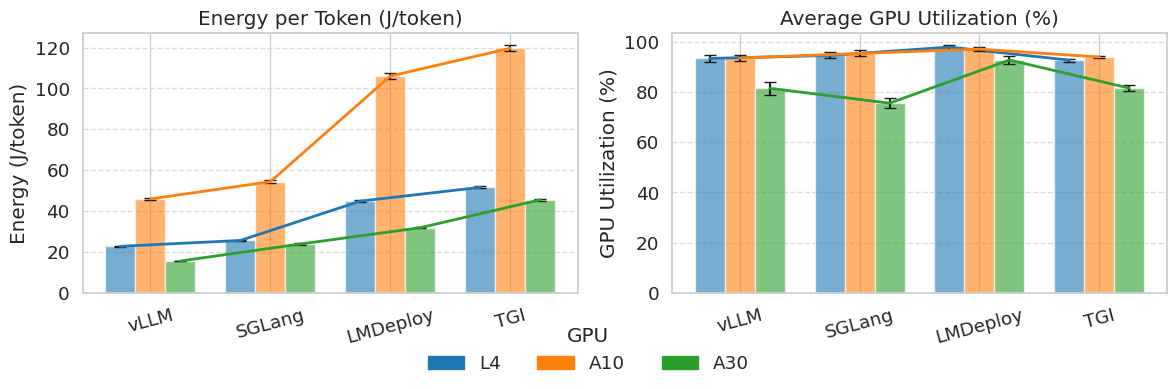}
  \caption{System-level runtime metrics across inference engines and GPUs (L4, A10, A30). 
  Bars show GPU utilization and energy consumption per token, with 95\% confidence intervals.}
  \label{fig:overall_system}
  \rmspace
\end{figure}

Figure~\ref{fig:overall_system} aggregates two central efficiency metrics, energy consumption and GPU utilization, across inference engines and GPUs. 
Results are averaged over three representative models, four tasks, and all three serving regimes to minimise model, task or serving-specific variance.  

The results show pronounced efficiency differences between engines. 
Energy per token and GPU utilization are not strongly correlated. 
vLLM consistently stands out as the most energy-efficient engine across GPUs, 
even though its average utilization is not always the highest. 
This suggests that its batching and scheduling strategies keep the GPU active 
only as long as necessary, thereby reducing wasted energy. 
In contrast, other engines may drive higher utilization but with less effective 
scheduling, leading to longer active periods and greater energy consumption. 
Hardware characteristics further amplify these effects: the A30, with its larger 
memory bandwidth and compute capacity, consistently reduces energy per token 
compared to the L4 and A10, widening the gap between efficient and less efficient 
engines.

\paragraph{Answering RQ3}
\emph{Efficiency depends as much on engine scheduling as on hardware}. High GPU 
utilization does not guarantee low energy cost: engines may keep devices 
busy without using resources effectively. vLLM shows that careful batching 
and scheduling can deliver the best energy efficiency even at moderate 
utilization. For practitioners, this means engine choice should consider 
how well GPU activity is converted into useful work. Hardware upgrades 
(e.g., L4 to A30) further reduce energy per token, but only when paired 
with engines that exploit the extra bandwidth and compute. Overall, 
vLLM provides the most balanced trade-off between energy efficiency and 
utilization.

\section{Conclusion}
\label{sec:conclusion}

We introduced Bench360, a novel framework for evaluating local LLM deployments by integrating task-specific quality with system metrics like latency, throughput, and energy consumption. Our work addresses the critical need for a holistic, user-focused benchmark for local inference.

Our results reveal that the optimal configuration is a complex trade-off. We found that while quantization enables larger, more accurate models on memory-constrained hardware, it often incurs significant latency and energy costs. Furthermore, the best-performing inference engine is highly scenario-dependent: LMDeploy excels at single-request responsiveness, SGLang and vLLM lead in batch throughput on different hardware tiers, and vLLM is the most robust and energy-efficient for multi-user serving.

Ultimately, our findings demonstrate there is no single best setup for local inference. The ideal choice demands a data-driven evaluation of the model, quantization, and engine against specific deployment requirements. Bench360 provides the NLP community with a vital tool to navigate this complex decision space and optimize deployments for their unique needs.

\clearpage
\section*{Limitations}
\label{sec:limitations}

While Bench360 enables a practical and deployment-oriented evaluation of local LLM inference, our study has several limitations that also point to directions for future work.

\emph{Quantization techniques.}
Our experiments focus primarily on GPTQ-based quantization.
Although Bench360 is designed to support additional quantization methods, broader coverage is currently constrained by inference-engine support.
Extending the evaluation to alternative techniques (e.g., activation-aware or mixed-precision methods) would provide a more comprehensive view as ecosystem support matures.

\emph{Model scale and hardware diversity.}
We evaluate models that fit within a 24\,GB VRAM budget on mid-tier GPUs, reflecting common local deployment environments in smaller labs, organizations, and individual settings.
While this choice ensures practical relevance, it limits conclusions about very large models and high-end accelerators.
Future work will extend Bench360 to larger models and a wider range of hardware configurations.

\emph{Inference engine coverage.}
Our comparison includes four inference engines, vLLM, SGLang, TGI, and LMDeploy, that support compatible model formats, Hugging Face Hub integration, and standardized APIs.
Other popular runtimes (e.g., llama.cpp/Ollama, ExLlamaV2, TensorRT-LLM) rely on specialized model formats or hardware-specific optimizations that hinder controlled comparison.
Bench360’s modular design allows such engines to be integrated once standardized interfaces or comparable execution paths become available.

\emph{Single-GPU focus.}
We restrict our evaluation to single-GPU deployments, reflecting common design choices for low-latency local inference where models are selected to fit entirely in device memory.
Multi-GPU inference and CPU offloading are not explored in this study, but are supported by the Bench360 architecture and represent important directions for future investigation.

\section*{Ethical considerations}
We do not recognize any ethical issues and conflicts with the ACL Code of Ethics. In fact, by making the trade-offs in local LLM inference more transparent, we are supporting the broader availability of local inference.

\paragraph{Energy and Carbon Footprint.}
The use of our benchmark may lead to a temporary increase in energy consumption for the evaluation of different LLMs and their deployment configuration. However, we believe this short-term cost is outweighted by the long-term benefits: Bench360 enables researchers and industry practitioners to make informed decisions about the LLMs to use in production environments, enabling them to find efficient solutions that reduce the energy consumption and its correlated CO2 emissions. 
The energy source for running our experiments has been based entirely on a 100\% renewable energy mix.

\paragraph{Responsible Deployment and Dual Use.} Beyond environmental impact, local LLM inference fundamentally alters the data privacy landscape. By keeping inference on-premise, organizations can process sensitive data (e.g., healthcare or legal records) without exposing it to third-party APIs, significantly enhancing privacy and data sovereignty. However, this decentralization introduces dual-use risks, as local deployments may lack the real-time safety filters and content moderation present in centralized API services. While Bench360 does not enforce safety guardrails, it contributes to risk mitigation by promoting transparency and auditability. By standardizing the evaluation of system metrics alongside task quality, our framework enables practitioners to design efficient, auditable local setups. This helps researchers better understand the trade-offs between a model's efficiency and its safety capabilities, fostering a more informed and responsible approach to open-weight deployment.

\paragraph{Use of Large Language Models (LLMs).} We have used LLMs to improve writing of this paper by drafting paragraphs of text, pasting them into cloud-based LLMs (Gemini, ChatGPT) for revision.

\bibliography{sources}

\clearpage
\appendix

\section{Bench360 Framework Details}

Table~\ref{tab:config} contains the configuration options of Bench360 with possible values that can be simply set by the user through a YAML configuration file.

  \begin{table*}[hb]
  \caption{YAML configuration fields with descriptions and possible values.}
  \centering
  \resizebox{1.0\hsize}{!}{
  \begin{tabular}{@{}lll@{}}
  \toprule
  \textbf{Field} & \textbf{Description} & \textbf{Possible Values} \\
  \midrule
  \texttt{hf\_model} & Hugging Face model identifier or path to the LLM & \texttt{[google/gemma-2-9b-it, ...]} \\
  \texttt{backend} & Inference engine for model execution & \texttt{[tgi, vllm, sglang, lmdeploy]} \\
  \texttt{quantization} & Precision format for model weights & \texttt{[fp16, int8, 4bit, awq, gptq]} \\
  \texttt{task} & Evaluation task type & \texttt{[mmlu, summarization, qa, sql]} \\
  \texttt{scenario} & Execution mode for inference workload & \texttt{[single, batch, server]} \\
  \texttt{samples} & Number of examples for batch/single scenarios & \texttt{[64, 128, 256, ...]} \\
  \texttt{batch\_size} & Batch size for batch processing scenario & \texttt{[8, 16, 32, ...]} \\
  \texttt{run\_time\_s} & Duration in seconds for server scenario & \texttt{[300, 600, 1200, ...]} \\
  \texttt{concurrent\_users} & Number of simultaneous users for server scenario & \texttt{[1, 8, 16, 32, ...]} \\
  \texttt{requests\_per\_user\_per\_min} & Request rate per user for server scenario & \texttt{[8, 16, 32, ...]} \\
  \bottomrule
  \end{tabular}
  }
  \label{tab:config}
  \rmspace
  \end{table*}

\section{System Performance Metrics}
\label{app:bench360-metrics}

Here we provide a more detailed and formal definition of the system performance metrics introduced in Table~\ref{tab:system_evaluation_metrics}.

\subsection{Latency}
We define three related latency measurements:  
\(\mathrm{TTFT}\) (time-to-first-token) is the delay between issuing a request at time \(t_{\mathrm{request}}\) and receiving the first token at \(t_{\mathrm{first\,token}}\);  
\(\mathrm{TPOT}\) (time per output token) is the total generation time \(T_{\mathrm{gen}}\) divided by the number of tokens \(N_t\);  
and \(\mathrm{GL}\) (generation latency) multiplies \(\mathrm{TPOT}\) by \(N_t\), which recovers the full \(T_{\mathrm{gen}}\):  
\begin{align}
\mathrm{TTFT} &= t_{\mathrm{first\,token}} - t_{\mathrm{request}}\\
\mathrm{TPOT}  &= \frac{T_{\mathrm{gen}}}{N_t}\\
\mathrm{GL}   &= \mathrm{TPOT} \times N_t = T_{\mathrm{gen}}
\end{align}

\subsection{Throughput}
We quantify generation speed in terms of tokens per second (TPS) and sentences per second (SPS). Here, \(N_t\) is the total number of tokens generated, \(N_s\) is the total number of sentences generated, and \(T_{\mathrm{gen}}\) is the generation time in seconds:
\begin{align}
\mathrm{TPS} &= \frac{N_t}{T_{\mathrm{gen}}}\\
\mathrm{SPS} &= \frac{N_s}{T_{\mathrm{gen}}}
\end{align}

\subsection{GPU Memory Usage}
We capture both average ($\overline{m}$) and peak ($m_{\max}$) GPU memory consumption over the measurement period. Here, \(m_i\) is the memory usage (MB) at sample \(i\), and \(n\) is the total number of samples:
\begin{align}
\overline{m} &= \frac{1}{n} \sum_{i=1}^n m_i\\
m_{\max}     &= \max_{1 \le i \le n} m_i
\end{align}

\subsection{GPU Utilization}
We capture both average ($\overline{u}$) and peak ($u_{\max} $) GPU utilization over the measurement period. Here, \(u_i\) is the GPU utilization percentage at sample \(i\), and \(n\) is the total number of samples:
\begin{align}
\overline{u}   &= \frac{1}{n} \sum_{i=1}^n u_i\\
u_{\max}       &= \max_{1\le i \le n} u_i
\end{align}

\section{System Cost \& Resource Metrics}

Here we provide a more detailed and formal definition of the system cost and resource metrics introduced in Table~\ref{tab:system_evaluation_metrics}.

\subsection{Power \& Energy}
This set of metrics defines average and peak power and derives energy consumption. Here, \(p_i\) is the power reading (W) at sample \(i\), \(n\) is the total number of samples, \(T_{\mathrm{gen}}\) is the generation time in seconds, \(N_t\) is the number of tokens generated, and \(N_s\) is the number of sentences generated:
\begin{align}
\overline{p}       &= \frac{1}{n} \sum_{i=1}^n p_i\\
p_{\max}           &= \max_{1\le i \le n} p_i\\
E_{\mathrm{Wh}}    &= \frac{\overline{p} \times T_{\mathrm{gen}}}{3600}\\
E_{\mathrm{J}}     &= E_{\mathrm{Wh}} \times 3600 = \overline{p} \times T_{\mathrm{gen}}\\
E_{\mathrm{token}} &= \frac{E_{\mathrm{J}}}{N_t}\\
E_{\mathrm{sentence}} &= \frac{E_{\mathrm{J}}}{N_s}
\end{align}

\subsection{Model Size \& Overhead}
 \(\mathrm{Overhead}\) captures the additional GPU memory consumed at runtime beyond the model’s static file size. Here, \(M_{\mathrm{size}}\) denotes the total size of the model’s weight files on disk (in MB), and \(\overline{m}\) is the average GPU memory footprint (in MB) measured during inference:
\begin{align}
\mathrm{Overhead} &= \max\bigl(\overline{m} - M_{\mathrm{size}},\,0\bigr)
\end{align}

\section{System Deployment \& Infrastructure Metrics}

Here we provide a more detailed and formal definition of the system deployment and infrastructure metrics introduced in Table~\ref{tab:system_evaluation_metrics}.

\subsection{Cold Start Time} \(T_{\mathrm{cold}}\) measures the total latency of a fresh inference invocation. Here, \(T_{\mathrm{startup}}\) denotes the time to initialize the inference framework, \(T_{\mathrm{load}}\) is the time to load the model weights into memory, and \(\mathrm{TTFT}\) (time‐to‐first‐token) is the latency from request issuance to receiving the first token:
\begin{align}
T_{\mathrm{cold}} &= T_{\mathrm{startup}} + T_{\mathrm{load}} + \mathrm{TTFT}
\end{align}

\section{Additional Experimental Details and Results}

\subsection{Decoding Parameters, Prompts and Reproducibility Details}

\paragraph{Decoding Parameters.} All experiments use fixed decoding parameters to ensure comparability across models, inference engines, and hardware platforms. Unless stated otherwise, text generation is performed with temperature $=0.1$ and top-$p = 0.9$. We set the maximum number of generated tokens to 64 for all tasks, except for summarization, where we set it to 128 output tokens to accommodate larger generation outputs.

\paragraph{Prompts.} All tasks are evaluated using \emph{few-shot prompting}. Task-specific prompt templates prepend a fixed set of demonstration examples before the evaluation input. These demonstrations are statically defined per task and remain identical across all models, inference engines, hardware configurations, and runs.

\paragraph{Reproducibility.} To ensure reproducibility, all sources of randomness are seeded with a fixed global seed. Specifically, we set \texttt{seed = 42} for Python’s \texttt{random} module, NumPy, and PyTorch.
Prompt templates, preprocessing logic, and task definitions are fully specified in the public repository (\githuburl) accompanying this paper. They can be inspected in the \path{benchmark/tasks/} folder seperately for each task, e.g., \path{benchmark/tasks/mmlu.py} contains the specification and prompt template for the MMLU task. For each task, evaluation samples are drawn uniformly without replacement, with sample counts controlled via the benchmark configuration and reported in the main text.

\subsection{Hardware Setup and Inference Frameworks}

Experiments were performed on three Kernel-based Virtual Machines (KVMs) running Ubuntu Linux. Table~\ref{tab:hardware-specs} summarizes the host system configurations, Table~\ref{tab:gpu_comparison} reports the architectural characteristics of the evaluated GPUs, and Table~\ref{tab:engine_comparison} compares the inference frameworks used in our study.
\vspace{1cm}

\begin{table*}[htbp]
  \centering
  \scalebox{0.75}{
    \begin{tabular}{@{}llll@{}}
      \toprule
      \textbf{Component}      & \textbf{Machine 1} & \textbf{Machine 2} & \textbf{Machine 3} \\ 
      \midrule
      CPU                     & AMD EPYC-Milan (1×16 cores)  
                               & Intel Broadwell (1×16 cores) 
                               & Intel Ice Lake (30 vCPUs) \\ 
      Memory                  & 47 GiB   
                               & 31 GiB 
                               & 200 GiB \\ 
      Storage                 & 500 GB NVMe 
                               & 812 GB NVMe
                               & 1.4 TB SSD \\ 
      GPU                     & NVIDIA L4, 24 GiB (CUDA 12.8)  
                               & NVIDIA A30, 24 GiB (CUDA 12.9) 
                               & NVIDIA A10, 24 GiB (CUDA 12.9) \\ 
      OS \& Kernel            & Ubuntu 6.8.0-58-generic  
                               & Ubuntu 5.15.0-139-generic
                               & Ubuntu 6.8.0-58-generic \\ 
      \bottomrule
    \end{tabular}
  }
    \caption{Hardware specifications of the three machines used in our experiments.}
  \label{tab:hardware-specs}
\end{table*}

\vspace{1cm}

\begin{table*}[htbp]
\centering
\footnotesize
\begin{tabular}{l c c c}
\toprule
\textbf{Spec} & \textbf{L4} & \textbf{A10} & \textbf{A30} \\
\midrule
GPU Architecture      & Ada Lovelace           & Ampere               & Ampere              \\
CUDA Cores            & 7,424                  & 9,216                & 10,752              \\
Memory                & 24 GB GDDR6            & 24 GB GDDR6          & 24 GB HBM2          \\
Memory Bandwidth      & $\sim$300 GB/s         & $\sim$600 GB/s       & $\sim$933 GB/s      \\
FP16 Tensor TFLOPs    & 242 TFLOPs*            & 125 / 250 TFLOPs*    & 165 / 330 TFLOPs*   \\
INT8 Tensor TOPS      & 485 TOPS*              & 250 / 500 TOPS*      & 330 / 661 TOPS*     \\
TDP                   & 72 W                   & 150 W                & 165 W               \\
BF16 Support          & Yes                    & Yes                  & Yes                 \\
INT4 GPTQ Support     & Yes                    & Yes                  & Yes                 \\
Form Factor           & Low-profile (PCIe)     & Full-height (PCIe)   & Dual-slot (PCIe)    \\
\bottomrule
\end{tabular}
\caption{Comparison of L4 \cite{NVIDIA_L4}, A10 \cite{NVIDIA_A10_datasheet}, and A30 \cite{NVIDIA_A30} GPUs for LLM inference. Values marked with * include structured sparsity (2:4).}
\label{tab:gpu_comparison}
\end{table*}

\vspace{1cm}

\begin{table*}[htbp]
\centering
\small
\begin{tabular}{l c c c c}
\toprule
\textbf{Feature} & \textbf{vLLM} & \textbf{SGLang} & \textbf{LMDeploy} & \textbf{TGI} \\
\midrule
Maintainer & UC Berkeley & LMSYS & OpenMMLab & Hugging Face \\
License & Apache 2.0 & Apache 2.0 & Apache 2.0 & Apache 2.0 \\
Quantization & FP16, GPTQ, AWQ & FP16, INT4 & GPTQ, AWQ & FP16, INT8 \\
KV Cache & PagedAttn & Fine-grained & Prefill-optim. & CUDA-based \\
Streaming & Yes & Yes & Yes & Yes \\
Multi-Model & No & Yes & Yes & Yes \\
Batching & Token-level & Static & Token-level & Token-level \\
API & OpenAI-comp. & OpenAI-comp. & OpenAI-comp. & OpenAI-comp. \\
Docker & Official image & Official image & Official image & Official image \\
\bottomrule
\end{tabular}
\caption{Comparison of local LLM inference engines used in our study: vLLM~\cite{vllm2023}, SGLang~\cite{sglang2024}, LMDeploy~\cite{lmdeploy2024}, and TGI~\cite{tgi2025}.}

\label{tab:engine_comparison}
\end{table*}

\begin{figure*}[htbp]
  \centering
  
  \begin{subfigure}{\textwidth}
    \centering
    \includegraphics[width=0.8\textwidth]{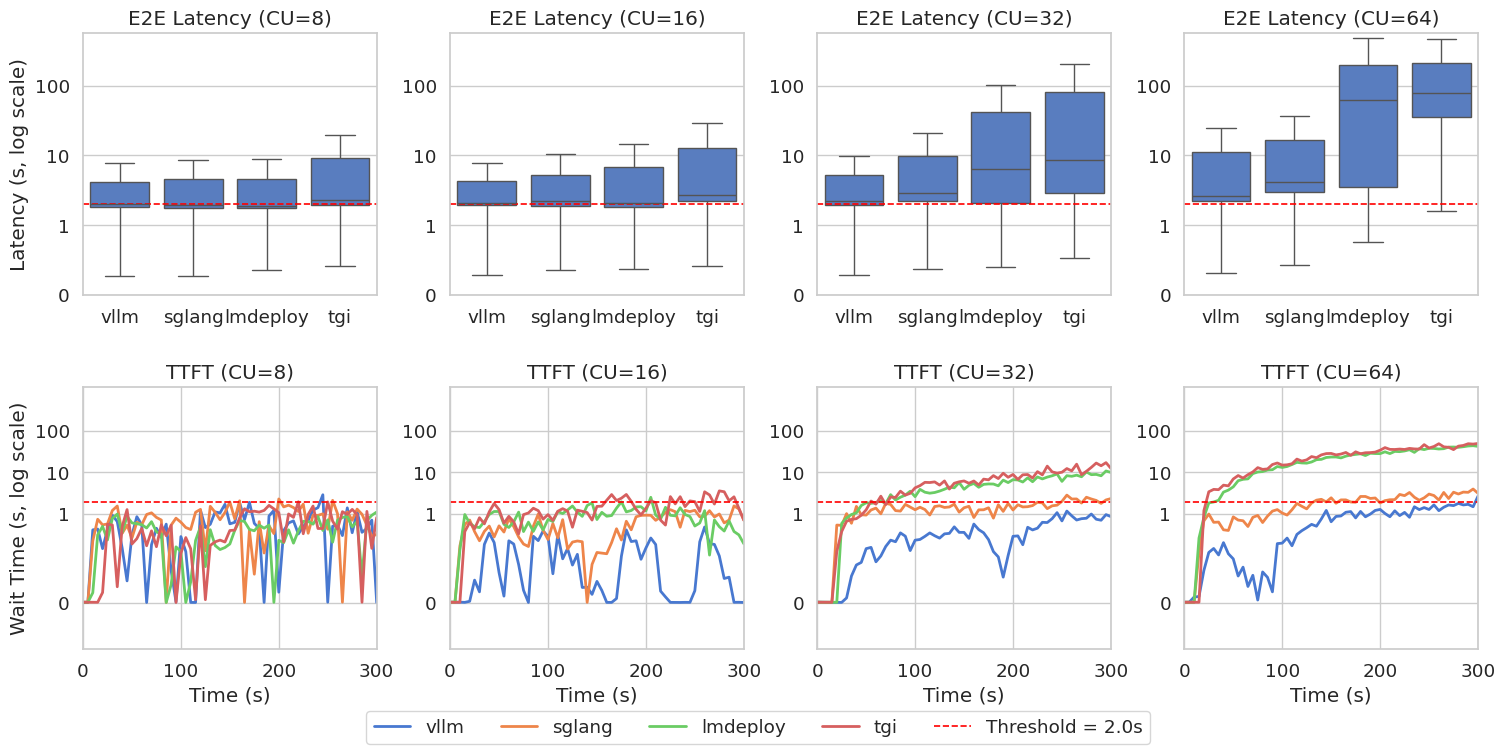}
    \caption{L4}
    \label{fig:l4}
  \end{subfigure}
  \vspace{1em} %

  \begin{subfigure}{\textwidth}
    \centering
    \includegraphics[width=0.8\textwidth]{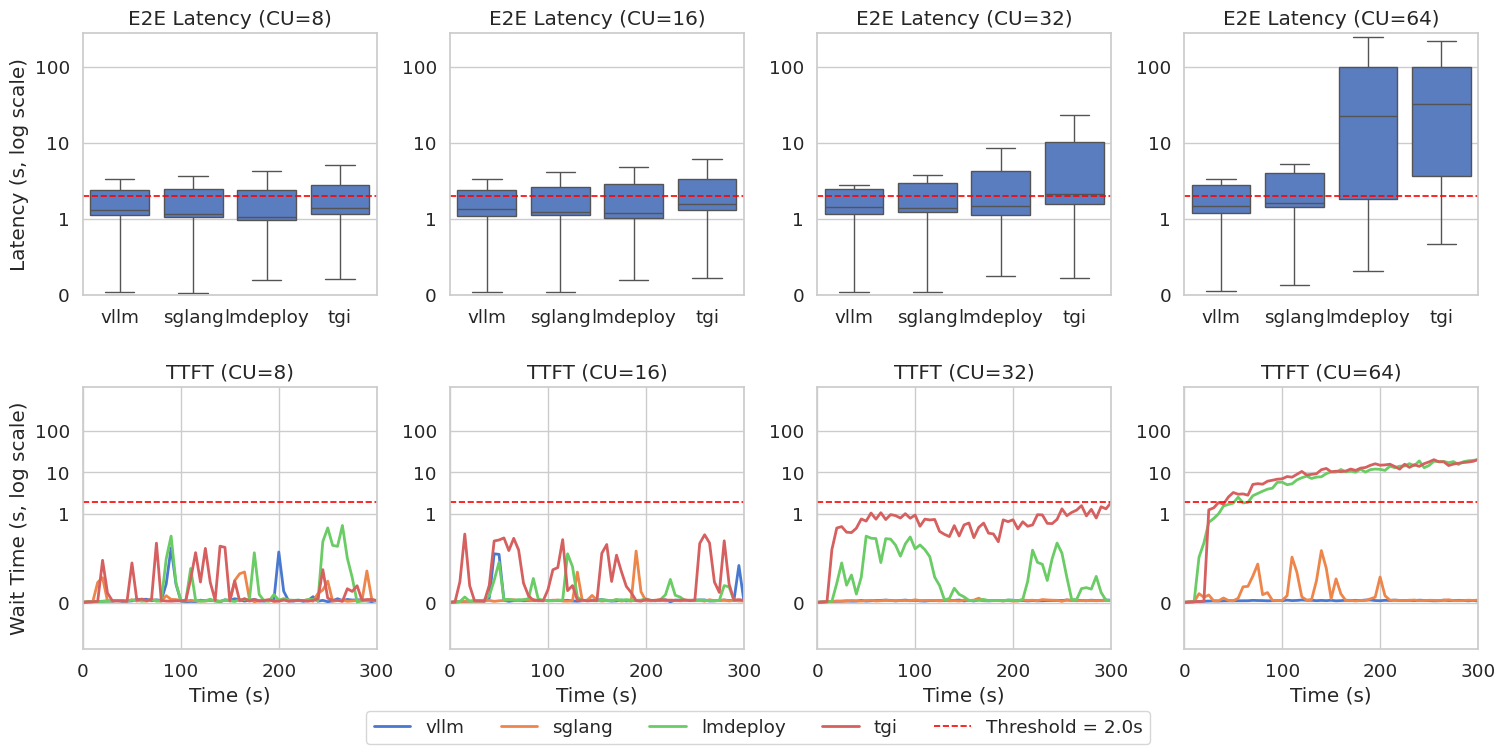}
    \caption{A10}
    \label{fig:a10}
  \end{subfigure}
  \vspace{1em} %

  \begin{subfigure}{\textwidth}
    \centering
    \includegraphics[width=0.8\textwidth]{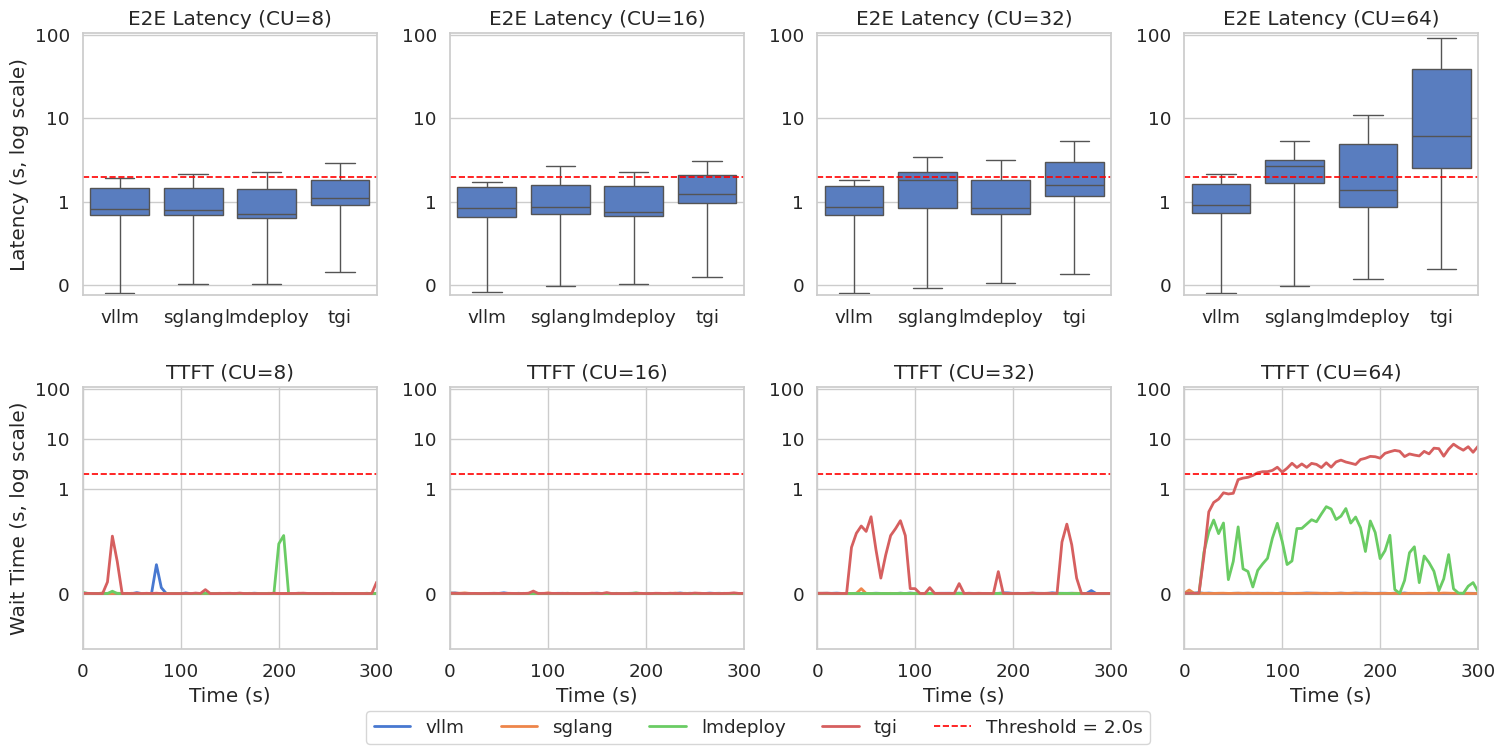}
    \caption{A30}
    \label{fig:a30}
  \end{subfigure}

  \caption{\textbf{Server-side concurrency across GPUs.} Rows in each panel show E2E (top, log scale) and TTFT (bottom) vs.\ concurrency. Red dashed lines indicate QoS thresholds: 2\,s TTFT and 6\,s E2E.}
  \label{fig:server_concurrency_all}
\end{figure*}

\end{document}